\documentclass[hidelinks]{ieeeaccess}

\usepackage{cite}
\usepackage{amsmath,amssymb,amsfonts}
\usepackage{algorithmic}
\usepackage{graphicx}
\usepackage{subcaption}
\usepackage{textcomp}
\usepackage{hyperref}

\usepackage{bm}
\usepackage{bbm}
\usepackage{adjustbox}
\usepackage{hhline}
\usepackage{booktabs}
\makeatletter
\AtBeginDocument{\DeclareMathVersion{bold}
\SetSymbolFont{operators}{bold}{T1}{times}{b}{n}
\SetSymbolFont{NewLetters}{bold}{T1}{times}{b}{it}
\SetMathAlphabet{\mathrm}{bold}{T1}{times}{b}{n}
\SetMathAlphabet{\mathit}{bold}{T1}{times}{b}{it}
\SetMathAlphabet{\mathbf}{bold}{T1}{times}{b}{n}
\SetMathAlphabet{\mathtt}{bold}{OT1}{pcr}{b}{n}
\SetSymbolFont{symbols}{bold}{OMS}{cmsy}{b}{n}
\renewcommand\boldmath{\@nomath\boldmath\mathversion{bold}}}
\makeatother

\def\BibTeX{{\rm B\kern-.05em{\sc i\kern-.025em b}\kern-.08em
    T\kern-.1667em\lower.7ex\hbox{E}\kern-.125emX}}

 \graphicspath{{./figures/}}

\hypersetup{colorlinks,urlcolor=blue}
\begin{document}
\history{Date of publication xxxx 00, 0000, date of current version xxxx 00, 0000.}
\doi{PREPRINT of Accepted Manuscript}

\title{Which Augmentation Should I Use? An Empirical Investigation of
  Augmentations for Self-Supervised Phonocardiogram Representation Learning}

\author{\uppercase{Aristotelis Ballas}\authorrefmark{1}, \uppercase{Vasileios
    Papapanagiotou}\authorrefmark{2}, \IEEEmembership{Member, IEEE}, and
  \uppercase{Christos Diou}\authorrefmark{1}, \IEEEmembership{Member, IEEE}}

\address[1]{Department of Informatics and Telematics, Harokopio University,
  Omirou 9, Tavros, 177 78 Athens, Greece (e-mail: <aballas, cdiou>@hua.gr)}
\address[2]{Department of Medicine, Huddinge, Karolinska Institutet,
  Blickagången 16, Flemingsberg, 141 52 Stockholm, Sweden (e-mail:
  vasileios.papapanagiotou@ki.se)}

\tfootnote{The work leading to these results has been funded by the European
  Union under Grant Agreement No. 101057821, project RELEVIUM.}

\markboth
{Ballas \headeretal: Which Augmentation Should I Use?}
{Ballas \headeretal: Which Augmentation Should I Use?}

\corresp{Corresponding author:Aristotelis Ballas (e-mail: aballas@hua.gr).}

\begin{abstract}
  Despite recent advancements in deep learning, its application in real-world
  medical settings, such as phonocardiogram (PCG) classification, remains
  limited. A significant barrier is the lack of high-quality annotated datasets,
  which hampers the development of robust, generalizable models that can perform
  well on newly collected, out-of-distribution (OOD) data. Self-Supervised
  Learning (SSL), particularly contrastive learning, has shown promise in
  mitigating the issue of data scarcity by leveraging unlabeled data to enhance
  model robustness and effectiveness. Even though SSL methods have been proposed
  and researched in other domains, works focusing on the impact of data
  augmentations on model robustness for PCG classification is limited. In
  particular, while augmentations are a key component in SSL, selecting the most
  suitable transformations during the training process is highly challenging and
  time-consuming. Improper augmentations can lead to substantial performance
  degradation, even hindering the network's ability to learn meaningful
  representations. Addressing this gap, our research aims to explore and
  evaluate a wide range of audio-based augmentations and uncover combinations
  that enhance SSL model performance in PCG classification. We conduct a
  comprehensive comparative analysis across multiple datasets and downstream
  tasks, assessing the impact of various augmentations on model performance and
  generalization. Our findings reveal that depending on the training
  distribution, augmentation choice significantly influences model robustness,
  with fully-supervised models experiencing up to a 32\% drop in effectiveness
  when applied to unseen data, while SSL models demonstrate greater resilience,
  losing only 10\% or even improving in some cases. This study also sheds light
  on the most promising and appropriate augmentations for robust PCG signal
  processing, by calculating their effect size on model training. These insights
  equip researchers and practitioners with valuable guidelines for building more
  robust, reliable models in PCG signal processing.

\end{abstract}


\titlepgskip=-21pt

\maketitle

\section{Introduction}
\label{sec:introduction}

Abnormal phonocardiogram (PCG) signals are often regarded as an important
indicator of heart diseases or deficiencies linked to increased mortality
\cite{gbd_2017_causes_of_death_collaborators_global_2018}. Since 
the collection of PCG signals is non-invasive, the automatic, accurate detection
of abnormalities in a sample has the potential to assist in early
prevention or timely treatment of such diseases, especially when access to
healthcare services for certain populations is limited
\cite{zuhlke2013congenital}.

The efficacy of automated methods for detecting abnormal audio samples, depends
to a large extent on the quality and nature of the audio recordings
themselves. Typically, signals are first manually analyzed and processed before
being used to train classifiers that discriminate between normal recordings and
samples that include indicators for different diseases \cite{khan2020}. However,
these classifiers often fail to produce generalizable and robust representations
of signals due to distribution shifts present in either the training data or on
the data that the model is evaluated on after training \cite{10233054}. For
example, a robust classifier should be able to maintain its accuracy on signals
that have been collected in different environments, from multiple devices, and
with varying sampling rates. Specifically for the case of PCG recordings,
additional factors such as microphone placement \cite{jaros2023novel,
  fontecave2019}, body size, and skin type \cite{giordano2021, huisa2023}, among
others, can significantly affect the signals. The above cases prove even more
challenging when screening an unborn child/fetus, where PCG can only be captured
through the mother's abdomen \cite{kahankova2023}. In addition to recording
conditions, the training of accurate ML models is dependent on annotated data,
the collection of which proves highly time- and cost-consuming. The above is
especially important in the biomedical domain, where available samples are
scarce to begin with.  As a result, classification models that have been trained
on data originating from a single source or on low-quality signals, fail to
generalize to previously unseen data distributions, which do not adhere to the
i.i.d. assumption (i.e., are not independently and identically distributed),
leading to degraded effectiveness \cite{recht2019imagenet, 10233054}.

A possible solution to the above limitations, which has proven advantageous in
various ML fields, is the utilization of unlabeled data \cite{9086055,
  10559458}. By training models on a large corpus of unlabeled data via
self-supervised learning (SSL), the feature extractor is able to learn general
representations and avoid overfitting on its training
distributions. Nonetheless, even though there have been works researching the
benefits of SSL in biosignal analysis, they are quite limited and primarily
focus on electrocardiography, electromyography or electroencephalography signals
(see Section \ref{sec:ssl4bio}). Therefore, in this work we turn our attention
to the Phonocardiogram domain and propose training robust classifiers for PCG
signal classification via contrastive self-supervised learning, leveraging both
labeled and unlabeled datasets. Furthermore, as the effectiveness of a backbone
encoder trained via SSL heavily depends on the adopted augmentation policy
\cite{Reed_2021_CVPR}, the design of a fitting augmentation policy for each
signal distribution is arguably the most critical and time-consuming step in an
SSL framework. To facilitate future research in this domain, we perform an
exhaustive evaluation on a large number of transformation combinations
(\textasciitilde$150$) and models (\textasciitilde$4,000$) applied on PCG signals and report their effect on
training and generalization. Specifically, we attempt to provide insight into
the following research questions:

\begin{itemize}
\item Does the application of contrastive SSL prove to be a valid training
  method for extracting robust and meaningful representations in the PCG
  classification domain?
\item If so, which augmentations or transformations lead to such representations
  and prove the most effective, and which actually inhibit training?
\item Is a classifier trained on the above representations able to maintain
  adequate effectiveness and demonstrate domain generalization capabilities when
  evaluated on OOD data, in contrast to its fully supervised counterpart?
\end{itemize}
To answer the above research questions, we develop and implement an extensive
comparative evaluation in which we: (a) investigate multiple combinations of
augmentations, leveraging both labeled and unlabeled datasets, (b) examine the
effectiveness on a global ``normal vs. abnormal'' downstream task, as well as
downstream tasks specific to each dataset (where applicable), (c) evaluate all
trained classifiers on OOD data that are left out during training, and finally,
(d) calculate the effect size of each augmentation.  To the best of our
knowledge, this is the first paper to research the effect of each applied
augmentation on learning robust OOD PCG representations.  To facilitate further
research in the field and provide a comprehensive baseline, we have open-sourced
our codebase\footnote{Code available at:
  \href{https://github.com/aristotelisballas/listen2yourheart}{https://github.com/aristotelisballas/listen2yourheart}}.

The rest of the paper is structured as follows. Initially, Section 
\ref{sec:relatedwork} presents previous work on PCG-based classification, SSL 
training, and model robustness. Subsequently, Section \ref{sec:methods} 
presents the methodology, including training approach, network architecture, 
augmentations, and datasets. Finally, Section \ref{sec:results_discussion} 
presents the evaluation framework and results, while Section 
\ref{sec:conclusions} concludes the paper.

\section{Related Work}
\label{sec:relatedwork}

Accurate and robust PCG classification has major implications in the well-being of patients as it can often be an indicator of heart 
deficiencies. However, the limited availability of annotated and quality data 
has hindered the progress of research in the field, along with biosignal 
classification in general. In this section, we focus on the most important 
papers and previous works in DL for PCG classification and in SSL for 1D 
biosignal processing. We also refer to methods which have been 
specifically developed for robust 1D biosignal classification.

\subsection{Deep Learning for Phonocardiogram Classification}
\label{dl4pcg}

The release of publicly available PCG datasets under the scope of the 2016
\cite{liu_open_2016} and 2022 \cite{reyna_heart_2023} PhysioNet
\cite{goldberger_physiobank_2000} challenges, has a sparked a line of research
in deep learning for PCG classification. The majority of previous works have
mainly proposed either 2D or 1D CNN architectures for tackling this particular
problem. The main difference lies in whether the proposed models are trained
directly on the one dimensional recorded auscultation waves or on spectral images
extracted from the audio recordings. For example, multiple recent papers
\cite{lu2022lightweight, 10081740, 10081866, 10081768} propose combining
mel-spectrogram images extracted from PCG signals with encoded patient
demographic data, in a dual-branch CNN model. The authors find that the
combination of the above representations leads to improved accuracy in the heart
murmur detection task. Similarly, in another work utilizing frequency domain
features, the authors of \cite{10081900} design a hierarchical deep CNN
architecture that extracts and combines features from multiple-scales of PCG
Mel-Spectrograms. A more recent work, \cite{10122565}, proposes training an
ensemble of 15 ResNets \cite{He_2016_CVPR} with channel-wise attention
mechanisms on similar spectrograms, for grading murmur intensity in PCG
recordings.

Although few, 1D models also prove promising in the heart-murmur detection
domain. In \cite{alkhodari_ensemble_2022} the authors employ a transformer-based
neural network and train it on 1D wavelet power features extracted from raw PCG
signals. On the other hand, \cite{10081735} proposes using a U-Net to predict
heart states from the raw PCG signal and combine the predictions with a ResNet
model for the final classification. In a different approach, the authors of
\cite{10081890} suggest that the combination of a 1D CNN-Bi-LSTM model coupled
with attention mechanisms, trained in conjunction with a feature-based classifier
leads to robust abnormal PCG signal detection.

In this work, we build upon our preliminary research
\cite{ballas_listen2yourheart_2022} for the 2022 PhysioNet Challenge. In
particular, we demonstrate the value of adopting self-supervised contrastive
learning for robust PCG classification and also report the top augmentations
that should be implemented in such a framework. Through our research, we aim to extensively investigate the benefits of each augmentation, and their combinations, for contrastive SSL in 1D PCG signal processing. The aim of this
study is to provide future researchers with best practices regarding PCG signal augmentation, while also proposing a structured protocol for the evaluation of developed methods.

\begin{figure*}[htbp]
	\centering
	\includegraphics[width=0.7\linewidth, ]{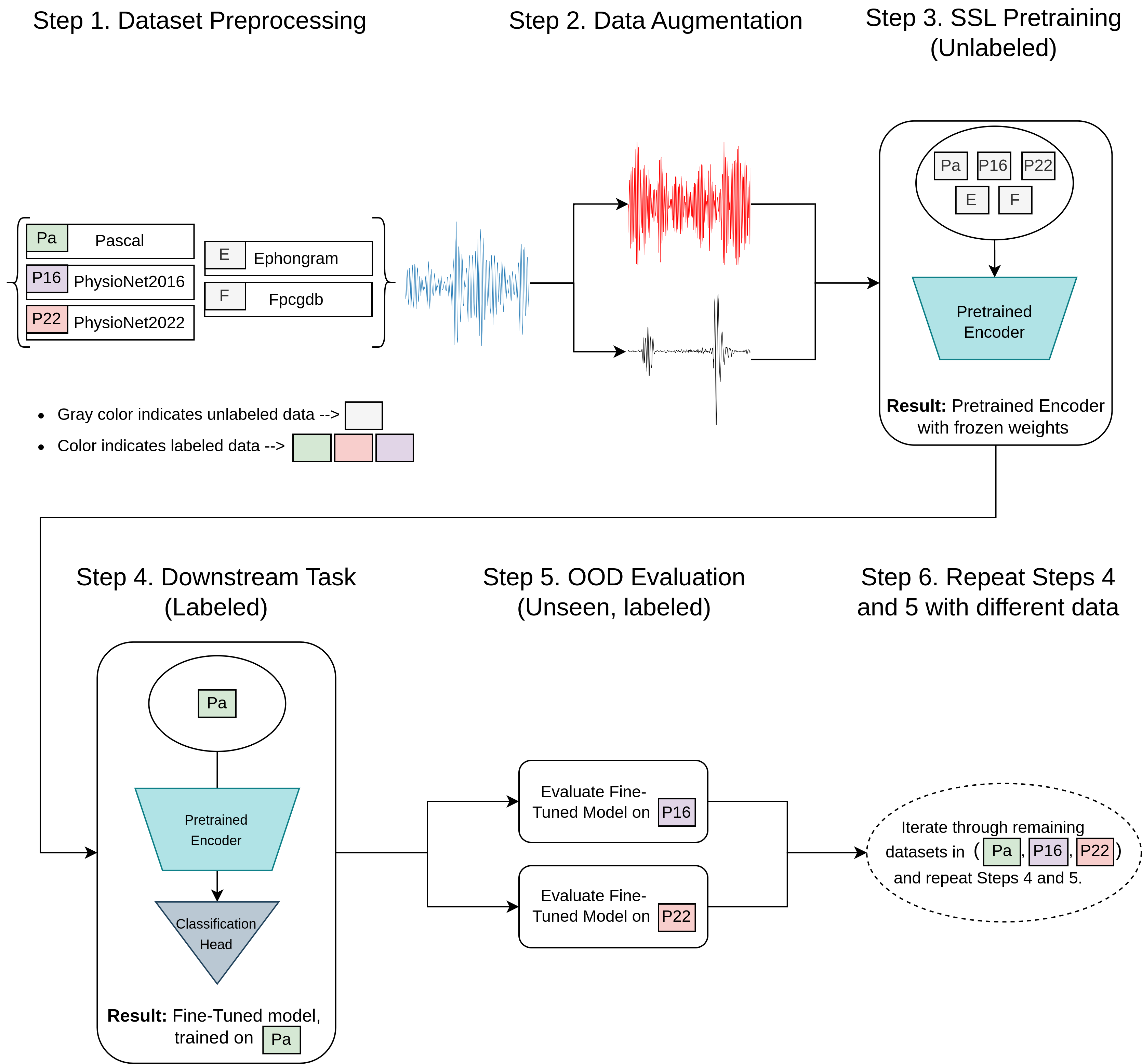}
	\caption{Illustration of the proposed experiment pipeline for training and
		evaluating the effectiveness and robustness of a model trained via
		Self-Supervised Contrastive Learning for PCG classification. The framework has six steps. In the first step, all datasets are prepared 
		and homogenized into a common format, as described in Section
		\ref{sec:ds_task}. In the second step, each signal is augmented to two
		versions. In the next step, the unlabeled and augmented signals are used to train the backbone encoder (Fig. \ref{fig:architecture} left) 
		to maximize the agreement between representations originating from the same initial signal. Following the pretraining phase, in step 4, a
		classification head (Fig. \ref{fig:architecture} right) is attached to the frozen weights of the pretrained encoder. The classifier 
		weights are fine-tuned on data drawn from one of the Pascal,
		PhysioNet2016 or PhysioNet2022 datasets, in a fully supervised 
		manner. The final model is then evaluated on the test split of the 
		same dataset. In the fifth step of the framework, the generalization 
		ability of the fine-tuned model is assessed on signals drawn from 
		datasets which are left-out ("unseen") during the training process in 
		step 4. We argue that given sufficient data and appropriate 
		augmentations, the pretrained encoder will be able to
		extract a generalized PCG representation regardless the dataset. 
		Finally, we repeat the fine-tuning and OOD evaluation steps for each remaining dataset (step 6), to assess the robustness of our method 
		when the classifier has been trained on different signal 
		distributions.}
	\label{fig:framework}
\end{figure*}

\subsection{Self-Supervised Learning in 1D biosignal classification}\label{sec:ssl4bio}

Self-supervised representation learning (SSL) \cite{9770283} was initially
proposed for mitigating the limit posed by the requirement of large annotated
datasets for practical and production-ready deep learning systems.  By taking
advantage of common features present in unlabeled data, SSL approaches rely on
pretext tasks \cite{misra2020self} to exploit them and ultimately provide models
which are able to extract generalized and robust representations. Inspired by
advancements in contrastive predictive coding \cite{oord_representation_2019},
the goal of Self-Supervised Contrastive Learning \cite{9462394,
  kumar_contrastive_2022} is to train a model to recognize augmented or
transformed views of the same initial signal. By distinguishing between
representations which originate from different original signals, the resulting
model is able to extract invariant representations from its input data and
reason upon features which remain stable across distinct data distributions
\cite{NEURIPS2021_fa14d4fe}. The SSL paradigm seems to fit quite well in the
medical and healthcare domain. In a field where sufficient annotated datasets
are scarcely available, there is a need to provide systems which are capable of
leveraging small amounts of data while yielding adequate and trustworthy
inference results. To this end, several works have proposed SSL approaches for
processing biosignals in the past, the majority of which have studied its
advantages when extracting representations from either electroencephalography
(EEG), electrocardiography (ECG), or electromyography (EMG) signals.

Regarding EEG signals, several prior studies 
revolve around developing methods for emotion recognition and/or sleep stage 
detection \cite{9837871}. The authors of \cite{mohsenvand_contrastive_2020} 
propose augmenting multi-channel EEG signals by recombining several channels 
into additional ones which represent different aspects of the EEG 
recording. On top of the combination, they augment each channel by applying 
predefined transformations and employ contrastive loss for improved 
emotion recognition and sleep-stage classification. In another work, the 
authors of \cite{banville_self-supervised_2019} argue that a model can learn 
informative EEG representations by predicting whether two windows of a signal 
are sampled from the same temporal context and use the above reasoning to 
design a relevant pretext task. The above study is extended in 
\cite{banville2021uncovering} where contrastive predictive coding is 
implemented on similar SSL-learned features to predict sleep stages and detect pathologies in multi-channel EEG signals. Finally, \cite{kostas_bendr_2021} proposed BENDR, which combines attention mechanisms
\cite{vaswani_attention_2017} and contrastive learning to produce a model with
increased generalization capabilities and fine-tune it for several downstream 
tasks in the EEG biosignal domain.

Several SSL methods have been proposed for ECG signal analysis as well
\cite{mehari_self-supervised_2022}, most of which focus on either emotion
recognition or pathology detection. For example, in
\cite{sarkar_self-supervised_2022} the authors argue that the training of a
model to recognize different ECG signal transformations can lead to generalized
and robust feature learning. They evaluate their method on four emotion
recognition datasets and report state-of-the-art results. Another paper
\cite{vazquez2022transformer} studying emotion recognition from ECG signals
introduces a transformer model which is pretrained to predict masked values of
the signal and then fine-tuned to predict the level of emotion in
individuals. Regarding the pathology detection task, most prior works focus on
discriminating normal ECG signals from others indicating a type of
arrhythmia. Among the self-supervised approaches proposed for ECG pathology
detection, contrastive learning is the most prevalent choice. Specifically,
authors either propose exploiting the temporal and spatial invariances of the
ECG signal \cite{kiyasseh_clocs_2021}, implementing attention mechanisms
\cite{oh_lead-agnostic_2022}, or combining wavelet transformations and random
crops of the signal \cite{chen_clecg_2021} for the pretext contrastive task,
before fine-tuning their resulting models for arrhythmia detection. In a
different approach, \cite{lan_intra-inter_2022} propose learning good
representations by contrasting signals both at the inter- and intra-subject
levels. Specifically, they argue that similarity should be maximized between
signals which share a common label but are measured from different individuals.

Finally, EMG signal processing is another interesting field that can 
greatly benefit from the application of robust prediction models 
\cite{9346072}. For example, \cite{liu_practical_2023} introduces NeuroPose, 
a system for 3D hand-pose tracking which aims to address challenges posed by varying sensor mounting or wrist positions, by pretraining
an encoder-decoder model to recognize augmented views of a single user's 
signal and then finetuning a downstream model to a subset of signals from a 
another user. Similarly, the authors of \cite{lai_contrastive_2022} employ 
a contrastive loss to pretrain a model on EMG signals for hand-gesture 
recognition.

\subsection{Model Robustness in 1D biosignal classification}\label{sec:robustbio}

Extracting generalized and robust representations has been one of the most
important long-standing goals of machine learning \cite{6472238}. To test and
improve the generalizabilty and robustness of deep learning models, several
fields have emerged, such as Domain Adaptation \cite{wang2018deep}, Transfer
Learning \cite{zhuang2020comprehensive}, and Domain Generalization
\cite{zhou2022domain, 10233054, ballas_multi}, which use data from non-IID
distributions or out-of-distribution (OOD) data for model evaluation. These
distinct data distributions are referred to as data \textit{Domains}, as they
are most often the outcome of different data-generating processes. Here, we
reference prior works which fall under the above research fields.

\textit{Domain adaptation} (DA) methods are the most popular choice when 
aiming to produce models with improved generalization capabilities in 1D biosignal classification. Similar to \textit{Transfer Learning} (TL), DA 
algorithms leverage the representation learning capabilities of models 
pretrained on diverse source data distributions and fine-tune them 
on previously unseen but similar target distributions. The 
end-result is a model which remains robust under the distribution shift 
between source and target data. In healthcare, DA methods are proposed for a 
plethora of biosignal classification tasks. In \cite{weimann_transfer_2021}, 
the authors propose pretraining vanilla CNN networks on a large corpus of 
raw ECG signals and demonstrate their effectiveness when fine-tuned on a 
smaller dataset for arrhythmia detection. In another more recent work, He et 
al. \cite{he_novel_2023} propose tackling the distribution shift between 
different arrhythmia detection ECG datasets, by employing three distribution 
alignment mechanisms. To be more specific, they first extract 
spatio-temporal features from preprocessed signals, pass them through a 
graph convolutional neural network and finally align the resulting 
representations between source and target domains. Many studies 
have researched the advantages of DA and TL in EEG classification as well 
\cite{wan_review_2021}. To name a few, the authors of \cite{10035017, 9154600} and \cite{peng_domain_2022} propose aligning source and target distributions via adversarial learning for motor image EEG classification and 
epilepsy detection respectively, while \cite{lan_domain_2018} investigates 
several state-of-the-art DA algorithms for EEG-based emotion recognition. 

Finally, \textit{Domain Generalization} (DG) algorithms aim to tackle 
model generalization at its core, as the main goal is to provide 
models that perform well across unseen datasets. Although limited, there has 
been a recent interest in biosignal DG research.
In the field of PCG classification, the authors of \cite{9298838} propose an 
ensemble classifier fusion method that yields improved results in a DG setup for 
abnormal vs normal signal classification. To tackle the domain shift and 
variability present in ECG signals collected from different patients, \cite{9344210} and \cite{ma_reducing_2019} propose domain-adversarial 
model training for ECG and EEG classification, respectively. Most recently,
the authors of \cite{10233054} introduce a DG benchmark and a model which
leverages representations from multiple layers of the network, inspired from 
\cite{9898255}, for 12-lead ECG and 64-channel EEG classification.

\begin{figure}[htbp]
	\centering
	\includegraphics[height=\columnwidth, ]{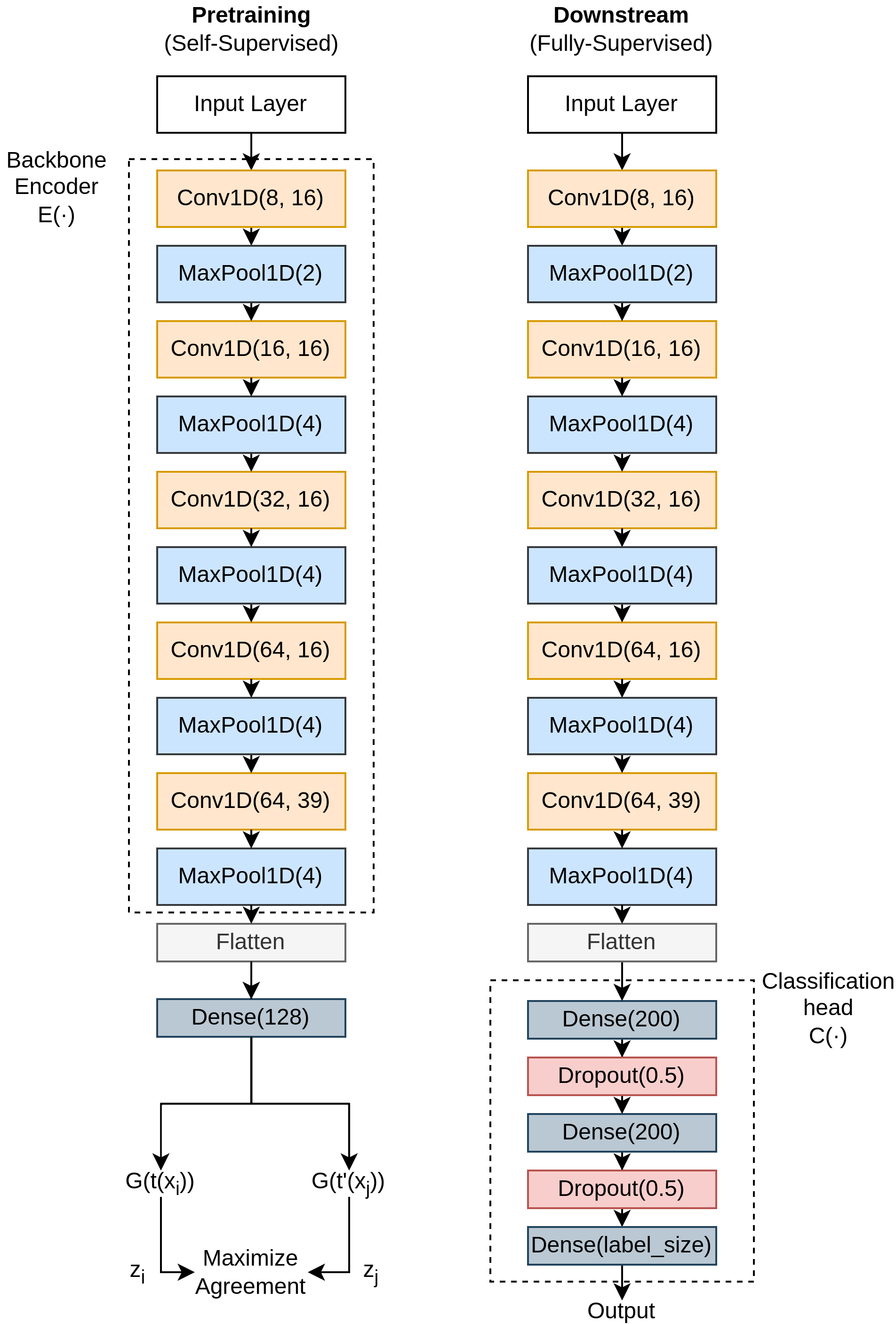}
	\caption{The architecture of the CNN encoder trained via SSL contrastive learning (left) and the classification head trained for the downstream task (right). The weights of the encoder or "backbone network" are frozen after the pretraining phase. During the downstream task, the classification head is attached to the pretrained encoder and its weights are fine-tuned on the dataset of said task in a fully-supervised manner. The architecture of the "downstream model" is used as a baseline in all of our experiments, where all of its layer weights are adjusted during fully-supervised training.}
	\label{fig:architecture}
\end{figure}

\section{Materials and Methods}
\label{sec:methods}

In this work, we introduce an evaluation framework for assessing the combination
of various transformations, as well as the robustness of a network trained via
contrastive SSL, against a fully supervised baseline model. This enables us to
evaluate and identify the most effective signal transformations or augmentations
within the PCG classification domain as well as highlight additional benefits
of training an end-to-end classification model via Self-Supervised Contrastive
Learning, such as improved model robustness and generalization.

For the self-supervised training of all models, we adapt the SimCLR
\cite{chen_simple_2020} methodology to 1D PCG representation learning and
evaluate numerous augmentation combinations designed specifically for 1D signal
processing tasks.

In the following sections, we describe the datasets and preprocessing steps
followed for all signals included in the study, provide a detailed description of the protocol for evaluating the significance of each combination of
augmentations, as well as the generalization ability of a model pretrained with
self-supervised contrastive learning. Finally, we present the architecture and
relevant implementation details of the 1D CNN network implemented for PCG
classification.

\subsection{Datasets \& Downstream tasks}\label{sec:ds_task}
The PCG signals used in this study originate from a total of 5 distinct public
datasets and are available as ``wav'' files. These include the Ephnogram
\cite{kazemnejad2021ephnogram}, FPCGDB \cite{cesarelli_simulation_2012}, Pascal
\cite{pascal-chsc-2011}, PhysioNet2016 \cite{liu_open_2016}, and PhysioNet2022
\cite{reyna_heart_2023} databases. Here, we describe each dataset separately and
provide all necessary details regarding the preprocessing steps followed to homogenize all signals into a common format.

The \textbf{Ephnogram} \cite{kazemnejad2021ephnogram} database contains a total
of 69 multi-channel, simultaneous ECG and PCG recordings from 24 healthy adults
aged between 23 and 29, acquired during physical activity and rest
conditions. Most recordings (62 during exercise) are 30 minutes long while the
duration of the remaining 7 (rest) is about 30 seconds. As specified by the
researchers, the dataset includes poor-quality sample signals making it appropriate for developing noise-resilient methods. As this dataset was proposed and gathered for multi-modal signal processing purposes and not for pathology detection or classification, it does not include labeled samples. For our study, we use the unlabeled PCG signals for the SSL pretraining of our models. All recordings were sampled at 8 kHz.

The Fetal PCG Database (FPCGDB) \cite{cesarelli_simulation_2012}, as its name suggests, is a dataset comprising 26 20-minute fetal phonocardiographic signals collected from healthy women during the final months of their pregnancies. The signals were sampled at 333 Hz. Similar to the Ephnogram dataset, these recordings were gathered for the development of fetal PCG signal simulation and fetal heart rate extraction algorithms, rather than for classification purposes, and thus do not include labeled samples. Additionally, many recordings are affected by noise, originating either from the pregnant body, such as muscular movements or placental blood turbulence, or from the surrounding environment.

The \textbf{Pascal} \cite{pascal-chsc-2011} Heart Sound Classification 
Challenge was introduced for developing algorithms for the automatic 
screening of cardiac pathologies. The challenge contains two tasks; one for 
heart sound segmentation and one for classification. In our work, we use the 
publicly available labeled data for model 
training and evaluation which was gathered in clinical and in-the-wild 
settings. Notably, the 832 PCG signals are classified as either Normal, 
Murmur, Extra Heart Sound, Artifact or Extrasystole\footnote{Each different 
	category is described in the official challenge description
	(\href{https://istethoscope.peterjbentley.com/heartchallenge/index.html}{https://istethoscope.peterjbentley.com/heartchallenge/index.html}).}. The 
recording lengths vary from 1 to 30 seconds and are sampled at 195 Hz. As each
sample is annotated with a single label, the above task is a 
multi-class classification problem.

The remaining two datasets were also introduced as publicly available
challenges for PCG classification. Similarly to Pascal, the
objective of the \textbf{PhysioNet2016} \cite{liu_open_2016} Challenge aims at
urging researchers to develop robust algorithms for differentiating between normal and abnormal heart-sound recordings as a binary classification problem. The data included for model training in the dataset contains a total of 3,126 PCG recordings collected from a single precordial location from both
healthy subjects and patients suffering from a variety of heart defections, lasting from 5 to around 120 seconds. Once again, the recordings were gathered 
in both clinical and non-clinical environments, with a sample rate of 2 kHz.

Finally, the \textbf{PhysioNet2022} Challenge \cite{reyna_heart_2023} was
designed for the development of algorithms that can detect heart murmur in
PCG signals. Building on the 2016 challenge, the 2022 edition contains
a diverse dataset of heart sound recordings collected from multiple auscultation
locations. The dataset includes 3,163 recordings obtained from 942 patients, with each patient contributing one or more PCG signals, recorded from one of the 
four heart valves or another prominent auscultation location. Each recording is classified into one of three categories: \textit{present} (heart murmur detected), \textit{absent} (no heart murmur detected), or \textit{unknown} (e.g., due to noise or corrupted audio), rendering this a multi-class classification task. All recordings were sampled at a rate of 4 kHz.

Each dataset contains signals of varying length, sampling rate and label
classes. To accommodate model training in our framework, we follow a similar
process to \cite{ballas_listen2yourheart_2022} for converting all signals
into a common format. As a first step and to avoid transient noise artifacts, we discard the first and final 2 seconds of each PCG recording. The
resulting signal is split into 5-second, overlapping windows with 50\%
overlap. After resampling all samples to 2 kHz, we assign two types of labels
to each window; one that follows the annotations of its original dataset (if it
is drawn from a labeled dataset) and one binary label indicating whether or not 
it derives from a normal heart-sound recording\footnote{In this study, all
recordings initially labeled as \textit{unknown} are considered abnormal.}.
For the SSL pretraining stage we assign pseudolabels to each window, as
described in Section \ref{sec:ssl}. The assignment of binary labels to each
window enables the evaluation of trained models across datasets, which do not
necessarily share similar sample classes. For example, a classifier trained to
detect normal PCG signals in the Pascal dataset, can also be evaluated without
further training on samples from the PhysioNet2022 dataset. The above
evaluation process is described in detail in Section \ref{eval_framework}.

\subsection{Self-Supervised Contrastive Learning for PCG Representation
	Learning}
\label{sec:ssl}

The main goal of this study is to first of all verify the use of SSL
contrastive learning, but also evaluate the importance of augmentation
selection for robust PCG representation learning. The proposed framework
consists of two separate tasks; a \textit{Pretraining} pretext task and a
\textit{Downstream} task, as depicted in Figure \ref{fig:framework}.

Assume a 1D PCG signal $x \in \mathbb{R}^{1 \times L}$ sampled from
$\mathcal{X}$, where $L$ is the length of the signal. The aim of the
contrastive SSL pretext task is to learn robust representations of $x$ from
unlabeled PCG signals, by maximizing the agreement between two different
augmented versions of itself (positive pair) and simultaneously minimizing the
agreement between entirely separate augmented windows. Essentially, a model is
trained to predict whether two augmented signals are the result of different
transformations applied on the original sample.  Formally, let $t(\cdot)$ be a
transformation or augmentation function drawn from a family of functions
$\mathcal{T}$. During the pretraining phase, an audio window $x$ is augmented
to two different versions $\tilde{x}$ and $\tilde{x}'$, after the application
of two transformation functions $t \sim T$ and $t' \sim T'$
respectively. Therefore, an initial batch of $N$ sample windows $(x_1,
x_2, \dotsc, \tilde{x}_n)$ results into an augmented batch of $2N$
signals $(\tilde{x}_1, \tilde{x}_2, \dotsc, \tilde{x}_n, \tilde{x}_1',
\tilde{x}_2', \dotsc, \tilde{x}_n')$, where $\tilde{x}_i = t(x_{i})$ and
$\tilde{x}_i' = t'(\tilde{x}_i)$ for $i \in [1, N]$. As mentioned, each
augmented pair of signals resulting from the same original window is considered
a positive pair, e.g $(\tilde{x}_1, \tilde{x}_1')$, while pairs of different
original windows comprise negative pairs, e.g $(\tilde{x}_1,
\tilde{x}_2')$. Following its augmentation, the $2N$ signal batch is passed
through a neural network encoder model $E(\cdot)$ (Figure
\ref{fig:architecture}), where each signal is encoded into a representation
$\tilde{h} = E(\tilde{x})$. 
Subsequently, the resulting representations are flattened and passed through a fully connected linear layer $G(\cdot)$ and projected into 128-dimensional vectors $\tilde{z}=G(\tilde{h})$, as proposed in \cite{chen2020simple}.

For the \textit{contrastive prediction} task, we employ the \textit{NT-Xent}
loss function introduced in \cite{oord_representation_2019} and implemented in
the SimCLR framework \cite{chen2020simple}. Specifically, given a set
$\{\tilde{x}_k\}$ which includes an augmented sample pair $\tilde{x}_i$ and
$\tilde{x}_j$, a model trained via contrastive SSL aims to identify
$\tilde{x}_j$ in $\{\tilde{x}_k\}$ for a given $\tilde{x}_i$. The
contrastive loss applied on the representations of any pair
$(\tilde{x}_i, \tilde{x}_j)$ is formulated as follows:

\begin{equation}
	\label{eq:contrastive_loss}
	\ell_{i,j} = -\log \frac{\exp(\mathrm{sim}(\tilde{z}_i, \tilde{z}_j)/\tau)}{\sum{k=1}^{2N} \mathbbm{1}_{k \neq i}\exp(\mathrm{sim}( \tilde{z}_i,  \tilde{z}_k)/\tau)},
\end{equation}
where $\tilde{z}_i$ and $\tilde{z}_j$ are the embedded representations of
$\tilde{x}_i$ and $\tilde{x}_j$ respectively, $N$ is the number of the initially
randomly sampled signals (before the augmentations, i.e.,
$\lVert {\tilde{x}_{k}} \rVert = 2N$), $\mathbbm{1} \in [0, 1]$ is an indicator
function and $\tau$ a temperature parameter. The $\mathrm{sim}(\cdot)$ function
calculates the cosine similarity between the encoded vectors and is defined as:
\begin{equation}
	\label{eq:cosine}
	\mathrm{sim}(\tilde{z}_i,\tilde{z}_j)=\frac{\tilde{z}_i.\tilde{z}_j}{\parallel \tilde{z}_i\parallel \cdot \parallel \tilde{z}_j\parallel}.
\end{equation}
The final loss is computed across all pairs in a mini-batch.

After the pretraining step has been completed (i.e., step 2 in Figure
\ref{fig:framework}), the weights of the backbone encoder network $E(\cdot)$
are frozen. At this point, the model is expected to produce a robust
representation $\tilde{h}$. While keeping the weights of the encoder frozen, a
classification head $C(\cdot)$, with randomly initialized weights, is attached
to the pretrained network. The layers of the classification head are then
trained on labeled audio windows $(x, y)$ in order to predict labels for each
downstream task. The classification layers are trained via fully supervised
learning, with the categorical Cross-Entropy (CE) loss. Depending on the labels
of each dataset in the downstream task, the CE loss is reduced to its binary
form.

\subsection{Augmentations for PCG Representation Learning}\label{sec:augmentations}
Arguably, the most integral part of the proposed framework is the family of
transformation functions $\mathcal{T}$, which is applied on each audio window, as the type
and intensity of each function can drastically affect the learning of the
encoder network. In order to conduct a rigorous evaluation of all augmentations
included in the proposed contrastive SSL framework and also report the functions
which should most likely be applied for robust PCG representation learning, we
select to implement functions that either simulate factors that can be found in
real world scenarios, such as the level (volume) of
noise. Additionally, the difference in sensor conductance can lead to different
scales of the recorded signals. To this end, it would make sense to at least
either add artificial noise to randomly scale each audio window during the
pretraining task.  Following the above rationale, in our study we select to
implement and evaluate the effectiveness of a total of 6 different
augmentations, each of which is described below:

\begin{itemize}
\item \textbf{Noise}: Zero-mean Gaussian noise is added to the PCG
  signal. Specifically, given noise
  $n(t)=\frac{1}{\sqrt{2\pi\sigma}}e^{-\frac{(x-\mu)^2}{2\sigma^2}}$, where
  $\mu$ is the mean and $\sigma$ the standard deviation, the resulting signal is
  $\tilde{x}(t) = x(t) + n(t)$.
	\item \textbf{Cut-off Filters}: Here either a high-pass or low-pass filter
is applied to the signal, at a 250, 500 or 750 Hz cutoff.
	\item \textbf{Scale}: The signal is randomly scaled by a factor of $a \in
[0.5, 2.0]$ to yield $\tilde{x}(t) = a \times x(t)$.
\item \textbf{Reverse}: The signal sequence is reversed, i.e.,
  $\tilde{x}(t) = x(l-t)$ where $l$ is the length of signal $x(t)$.
	\item \textbf{Inversion}: As its name suggests, this transformation inverts
the original signal by multiplying it by -1, i.e., $\tilde{x}(t) = - x(t)$.
	\item \textbf{Random Flip}: This function takes as input the signal and
concurrently applies the `reverse' and `inversion' augmentations based on a
given probability ($p$) between 0.3, 0.5 and 0.7.
\end{itemize}

In the proposed framework, we also evaluate the composition of augmentations 
by applying one transformation after the other. Specifically, we consider 4
different augmentation cases: 0vs1, 1vs1, 1vs2 and 2vs2. In each case, the 
signal is either not transformed, transformed once or augmented by the 
application of 2 concurrent functions.

\begin{figure*}[t]
	\centering
	\includegraphics[width=\linewidth, ]{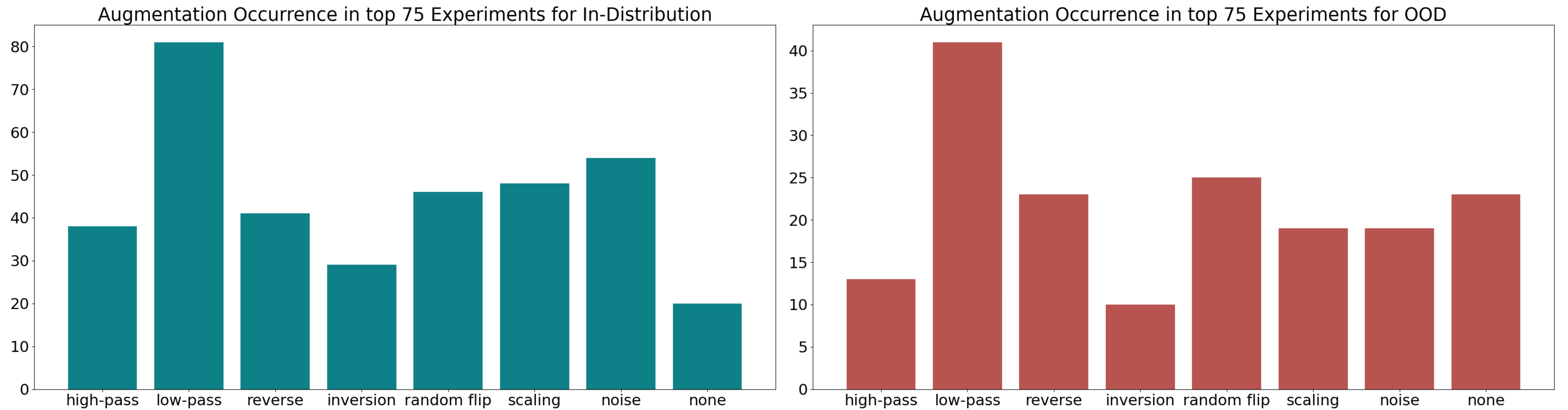}
	\caption{Number of occurrences of each augmentation type in the top 75
		performing models in each downstream task, leading to 150 total augmentation pairs. This offers and indication on
		the importance of each augmentation across all experiments. The plot on
		the left presents the result for In-Distribution experiments, while the
		plot on the right the results for the OOD experiments.}
	\label{fig:augmentation}
\end{figure*}

\subsection{Evaluation Protocol}\label{eval_framework}

The present study aims at providing future researchers with a roadmap
for applying SSL in Phonocardiogram processing problems. To this end, we 
design a rigorous and comprehensive protocol for evaluating the importance
of each applied augmentation in robust PCG classification. Each different 
augmentation combination is evaluated by repeating the experiment illustrated in Figure \ref{fig:framework}, with the exact same data splits for fair 
comparison. 

In the first step of the framework, we create a pool of 5-second,  unlabeled 
audio windows from all available datasets, namely Pascal, PhysioNet2016, 
PhysioNet2022, Ephnogram and FPCGDB. After the preprocessing step has been 
completed and all signals are homogenized in the common format described in 
Section \ref{sec:ds_task}, they are randomly shuffled and split into training and validation splits (80\% and 20\% respectively). In the second step, the 
split signals are cut into batches. Each sample in the batch is then 
duplicated and augmented by applying either one or a combination of the 
functions described in Section \ref{sec:augmentations}. The augmented samples
are then passed through a Neural Network encoder, which is trained
via the contrastive pretraining task (Step 
3). After the SSL backbone model has been trained, its weights are frozen.

In Step 4, a classification head with learnable weights is attached to the
pretrained encoder which is then trained on data from one of the labeled
datasets (Pascal, PhysioNet2016 or PhysioNet2022). Specifically, each of these
datasets is split into training, validation and test splits with a standard
70\%-20\%-10\% ratio and the trained model is evaluated on the test split
only. Since each model is evaluated on test data originating from the same
distribution as its training data, we call this the "In-Distribution"
evaluation. Inspired by recent work in biosignal domain generalization
\cite{10233054}, we also design an evaluation protocol for assessing the
robustness of a trained classifier on out-of-distribution data, which are drawn
from entirely different datasets. Since the data has been completely left-out
during downstream training, the above process is followed for evaluating the
generalization ability of a model. Specifically, we now also freeze the weights
of the attached classification layers and test the model on data drawn from the
remaining 2 labeled datasets. For example, if the network is trained for the
Pascal downstream task, it is evaluated on both in-distribution data (Pascal)
and on out-of-distribution data (PhysioNet2016 and PhysioNet2022). We call this
step the out-of-distribution or OOD evaluation step. However, in order to
evaluate the same model on each separate dataset, all data must share common
labels. To that end, we implement two different downstream task types: `All'
and `Binary'. In the `All' downstream task, the original labels of each dataset
are assigned to each audio window, while in the `Binary' task a signal is
labeled as either `normal' or `abnormal', as described in Section
\ref{sec:ds_task}. In the case of the PhysioNet2016 dataset, the two tasks
match as it introduces a binary classification problem in the first place. For
a fair comparison, all test data splits remain the same across each evaluation
scheme.  Finally, to assess the pretrained backbone encoder (and therefore the
applied augmentations) on all available data, we repeat Steps 4 and 5 with the
remaining datasets in a round-robin fashion (step 6 of the proposed framework). The significance of each augmentation and transformation is evaluated based on: (a) their occurence frequency in the top performing models in both In-Distribution and OOD evaluation splits, and (b) their 
effect on model training, as calculated by Cohen's d measure \cite{cohen2013statistical}. Additional information and details regarding augmentation evaluation are provided in Section \ref{sec:aug_eval}.

To thoroughly evaluate the proposed method we also repeat the entire experiment
by leaving data out during Step 1 and pretraining the model with fewer
data. This is also a means to concretely evaluate the proposed method on true
OOD data, as they are completely hidden throughout the training process, either
pretraining or downstream training. The best augmentation combinations are
selected based on the effectiveness of the downstream models on the
`In-Distribution' test split. The same models are then evaluated on the hidden
`OOD data'.

\subsection{Implementation Details}\label{implementation_dets}
In this section, we describe all the implementation details regarding the 
execution of our experiments. First of all, for the backbone encoder model
we implement the same model used in \cite{ballas_listen2yourheart_2022} and 
\cite{papapanagiotou_chewing_2017}. The encoder model is a 5 layer 
convolutional neural network, the architecture of which is depicted in
Figure \ref{fig:architecture} (left). Given a batch size of $b$ the input of the encoder model is a tensor of dimension $(b, 10000)$.

For the contrastive SSL pretraining phase, we train the convolutional layers of the network by projecting the flattened output of the last layer, with an original dimension of $(b, 8, 64)$ into a 128-D space via a dense or fully-connected layer and by ultimately minimizing the NT-Xent loss (eq. \ref{eq:contrastive_loss}). The dimension of the projection was set to follow the one in the original SimCLR paper \cite{chen2020simple}. For the cosine similarity (eq. \ref{eq:cosine}) we use a temperature of 0.1, as in \cite{ballas_listen2yourheart_2022}, while the initial batch size is set to 256 (leading to 512 after augmentation occurs). The SSL training epochs are set to a maximum of 200 epochs, based on an early stopping set on the validation split with a patience of 10 epochs. Additionally, to accommodate the large batch size we employ the LARS optimizer \cite{ginsburg2018large} with a linear learning rate warm-up of 20 epochs, up to 0.1, followed by a cosine decay of 1\%. 

During the downstream task, the dense projection layer of the pretrained 
encoder is discarded and all convolutional weights are frozen. Followingly, a 3-layered classification head is appended to the pretrained network, as shown
in Figure \ref{fig:architecture} (right). The dimensions of each layer are set to reflect the model implemented in \cite{ballas_listen2yourheart_2022}, which was shown to yield promising results. Dropout layers are included between each of the dense layers of the classification head in order to avoid overfitting. The application of dropout between multiple fully connected layers has shown improve the generalizability and robustness of classifiers \cite{srivastava2014dropout,papapanagiotou_chewing_2017}. The output size of the last decision layer depends on the downstream task type and the label size in each dataset. For the `Binary' type and the PhysioNet2016 dataset, the output size is set to 1. For the `All' type, the output size is set to 5 or 3, for the Pascal and PhysioNet2022 datasets respectively. In all cases, the classification head is trained via the appropriate cross-entropy loss and the Adam optimizer. Additionally, the learning rate is set to $10^{-4}$ and the batch size is set to 32. Once again, we train the model for 100 epochs at most, using early-stopping with 20 epochs patience. The same CNN model is 
used for the fully-supervised baseline.
All experiments are implemented in Tensorflow 2.12 
\cite{tensorflow2015-whitepaper} and trained on a SLURM \cite{yoo_slurm_2003} 
cluster, containing 4$\times$40GB NVIDIA A100 GPU cards, split into 8 20GB
virtual MIG devices.

\section{Results and Discussion}
\label{sec:results_discussion}

\subsection{Augmentation and Transformation Evaluation}\label{sec:aug_eval}

\begin{table}\centering
	\begin{center}
		\caption{The effect size of each augmentation on the SSL contrastive
			training. The effect were measured based on the out-of-distribution micro F1 scores for the PhysioNet2016 dataset, when trained on signals from PhysioNet2022, sorted by effect size. LP and HP indicate Low and High Pass filters, respectively.}
		\label{tab:effect_size}
		\begin{tabular}{c||c}
			\hline\noalign{\smallskip}
			\textbf{Augmentation} &  \textbf{Cohen's $d$} ($\uparrow$) \\
			\noalign{\smallskip}
			\hline 
			\noalign{\smallskip}
			
			Random Flip (0.5)       & 1.8012  \\
			Cut-off (500, 450) (LP) & 1.3295  \\
			Cut-off (750, 700) (LP) & 1.2839  \\
			Cut-off (250, 200) (LP) & 1.2639  \\
			Random Flip (0.3)       & 0.5784  \\
			Random Flip (0.7)       & 0.4923  \\
			Scale (1.0, 1.5)        & 0.2068  \\
			Noise (-0.01, 0.01)     & 0.1159  \\
			Inversion               & -0.0008 \\
			Reverse                 & -0.0039 \\
			Scale (1.5, 2.0)        & -0.1169 \\
			Cut-off (500, 550) (HP) & -0.1149 \\
			Cut-off (250, 300) (HP) & -0.2079 \\
			Noise (-0.001, 0.001)   & -0.2398 \\
			Noise (-0.1, 0.1)       & -0.3257 \\
			Scale (0.5, 2.0)        & -0.4016 \\
			Cut-off (750, 800) (HP) & -0.4743 \\
			
			\noalign{\smallskip}
			\hline
		\end{tabular}
	\end{center}
\end{table}

To avoid the combinatorial explosion poised by the evaluation of all possible
augmentation combinations, we designed a protocol to limit the experiments to a
feasible amount. As mentioned in Section \ref{sec:augmentations}, there are a
total of 7 (including not transforming) possible functions that can be applied
to the two copies of the audio sample, resulting in 4 possible augmentation
cases; 0vs1, 1vs1, 1vs2 and 2vs2. As a first step, we examined all
possible combinations in the 0vs1 and 1vs1 cases without applying the same
transformation twice. After evaluating the above we selected a subset of
augmentations which led to the top performing models in the ``In-Distribution'' setting and continued experimenting with 1vs2 and 2vs2 augmentation schemes. The models with the best results are presented in Table \ref{tab:experimental_results_ood} and are discussed in the next section.

The main aim of this study is to provide an evaluation of the most
appropriate augmentations for PCG signal analysis. Due to the fact that by only presenting the combination of the augmentations leading to the best
classification results could be misleading (since only the specific 
combination could pertain to said results), we present a plot depicting the 
frequency of augmentation occurrence in the top 75 experiments across all 3 
downstream tasks and for both In-Distribution and OOD data. Specifically, we select the top 25 results for each task and present the occurrences for all 75 total experiments in Fig. \ref{fig:augmentation}. The evaluation results 
are quite interesting as several conclusions can be drawn. First of all, the application of low-pass filters seem to be dominant in both In-Distribution 
and OOD top experiments. This is a clear indicator that such filters should 
be applied towards increasing the robustness of a classifier. What is more, the application of uniform noise, random flip and scaling on the signal, also 
boosts the effectiveness of classifiers in all cases. Furthermore, we found 
that the best performing models in each of the 3 downstream tasks, had 
adopted the following augmentations during their pretraining:

\begin{itemize}
	\item Pascal: low pass filter (pass band 250Hz, stop band 300 Hz) and random flip (p=0.7) vs inversion and uniform noise (mean 0, std 0.1), 
	\item PhysioNet2016: low pass filter (pass band 250Hz, stop band 300 Hz) vs reverse
	\item PhysioNet2022: low pass filter (pass band 250Hz, stop band 300 Hz) vs random flip (p=0.7) and random scaling with $a$ between 1.0 and 1.5	
\end{itemize}

While the above augmentations agree, for the most part, with the distributions
and number of occurrences in Fig. \ref{fig:augmentation}, they also support the
findings in several previous contrastive SSL works
\cite{chen_simple_2020,gopal_3kg_2021}, in which stronger augmentations lead to improved results. 

To further validate our results and to draw concrete conclusions regarding the
importance of each applied transformation, we calculate Cohen's $d$ 
\cite{cohen2013statistical} for each implemented function. Cohen's $d$ 
provides a standardized measure of the effect size of each augmentation and 
allows the comparison of their magnitude. Given two groups $x_1$ and 
$x_2$, Cohen's $d$ is the difference in their means divided by the pooled standard deviation ($s$). Specifically, Cohen's $d$ is defined as:  

\begin{equation}
	d = \frac{\bar{x_1} - \bar{x_2}}{s}
\end{equation}

where the pooled standard deviation is defined as:

\begin{equation}
	s = \sqrt{\frac{(n_1 - 1) {s^2}_1 + (n_2 - 1) {s^2}_2}{n_1 + n_2 - 2}}
\end{equation}
and $n_1$, $n_2$ are the variances for each group respectively. The consensus on the interpretation of Cohen's d is generally that its sign means to 
either a negative or positive effect, while a magnitude of $d = 0.01$ up to $d = 2.0$ corresponds to a very small or very large effect, respectively. To 
calculate Cohen's $d$, we create two subgroups for each augmentation. 
Specifically, let $a$ be an augmentation under consideration. The first group 
($x_1$) contains all experiments where $a$ was implemented, while the second 
group ($x_2$) consists of all experiments with the same set of augmentations 
but without $a$. For example, say that the augmentation under consideration is \emph{Noise}. One experiment to include in the first group would be the \emph{Noise vs Inversion} experiment. Therefore the corresponding experiment 
which should be added to the second group would be \emph{None vs Inversion}. 
The same process can be repeated for all augmentations and for all runs, 
including $2vs2$ experiments, as specified in Sec. \ref{sec:augmentations}. 
Accordingly, we calculate the effect of each augmentation based on the 
Out-of-Distribution Micro F1 scores for the PhysioNet2016 dataset, when 
trained on signals from PhysioNet2022. The magnitude of effect regarding each augmentation is presented in Table \ref{tab:effect_size}. 

Interestingly enough, the statistics presented in Fig. \ref*{fig:augmentation}, seem to agree with the effect sizes of each augmentation (Table \ref{tab:effect_size}). Based on Cohen's $d$ 
measure, the low-pass filters have the most important positive effect on 
robust SSL PCG learning. In simple terms, this means that the addition or 
application of low-pass cut-off filters during the pretraining phase of the encoder plays a significant role ($d > 1.0$) in learning robust 
representations. The above result agrees with the general consensus that important biosignal artifacts are present in low frequencies, whearas unrelated noise is attributed to higher frequencies \cite{bailey1990recommendations, SORNMO200555, vu2023importance}. Additionally, the application of \emph{Reverse} and \emph{Inversion} (i.e Random Flip) has an overall positive effect on training, as shown by the positive $d$ measure. What's more, the balanced reversion of inversion of the signals (i.e Random Flip with a .5\% chance), has the largest positive effect ($d \approx 1.8$), whereas the application of only one of the above flip augmentations seems to have approximately no effect on training. Similarly, but to a less degree, scaling the signals by a small factor, assists the model in extracting meaningful representations. On the other hand, the addition of noise, the application of high-pass filters and large scaling, generally appear to have a negative effect. The intuition behind the negative effect of noise, is that since the selected signals derive from an already ``noisy'' dataset, the additive noise renders them meaningless, while the 
application of high-pass filters removes important signal artifacts present 
in lower frequencies. 

Even though low-pass filters seem to be have the largest postive effect in 
training, they have to be combined with further transformations for a SSL 
model to reach optimal performance, as already mentioned. In the next section we present numerical results for all downstream tasks and experiments.

\begin{figure}[!t]
	\centering
	\begin{subfigure}[b]{0.45\columnwidth}
		\centering
		\includegraphics[width=\textwidth]{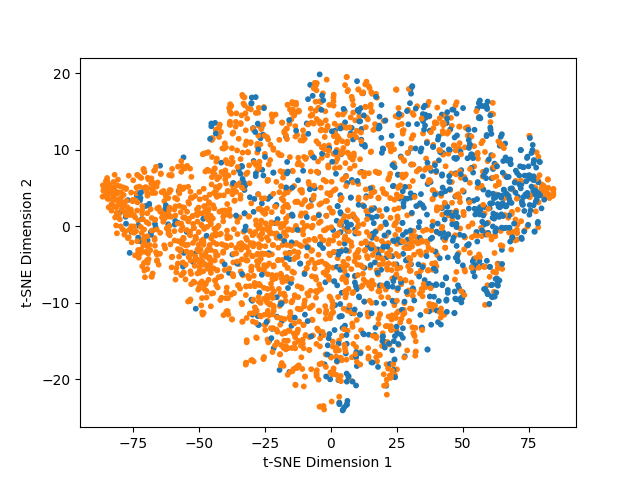}
		\caption{P16 - Baseline}
		\label{fig:tsne-baseline}
	\end{subfigure}
	\hfill
	\begin{subfigure}[b]{0.45\columnwidth}
		\centering
		\includegraphics[width=\textwidth]{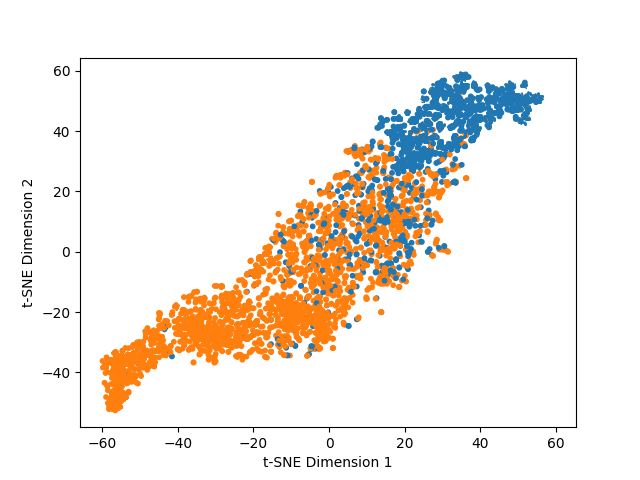}
		\caption{P16 - SSL}
		\label{fig:tsne-ssl}
	\end{subfigure}
	
	\vskip\baselineskip
	
	\begin{subfigure}[b]{0.45\columnwidth}
		\centering
		\includegraphics[width=\textwidth]{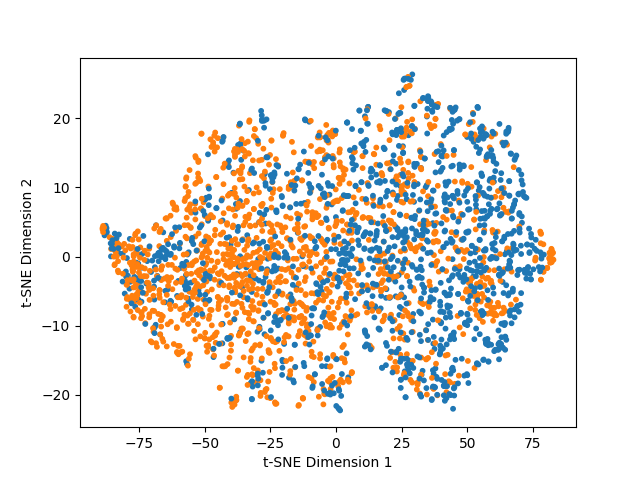}
		\caption{P22 (OOD) - Baseline}
		\label{fig:tsne-ood-base}
	\end{subfigure}
	\hfill
	\begin{subfigure}[b]{0.45\columnwidth}
		\centering
		\includegraphics[width=\textwidth]{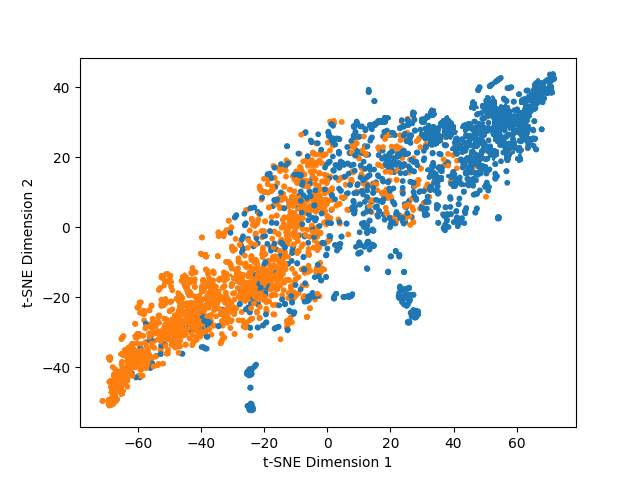}
		\caption{P22 (OOD) - SSL}
		\label{fig:tsne-ood-ssl}
	\end{subfigure}
	
	\caption{T-SNE visualizations of feature vectors extracted from the penultimate layer, after model training of a baseline fully-supervised model and a model trained via contrastive SSL. In order to exhibit the generalization capabilities of a model trained via the proposed framework, we plot the features of both In-Distribution, from the PhysioNet2016 dataset and OOD samples from the PhysioNet2022. Abnormal signals are colored in blue, while normal signals are shown in orange. As illustrated, the features extracted by the baseline models are chaotic and perplexed, whereas the SSL model seems to group samples in a more structured manner.}
	\label{fig:tsne}
\end{figure}

\subsection{Results}\label{sec:results}

\begin{table*}\centering
	\caption{This table presents the results of the best performing models (as reported in Section \ref*{sec:aug_eval})Top results are specified in bold face. The
		first column of the table indicates the datasets on which the SSL backbone
		was pretrained on. E, F, Pa, P16 and P22,
		indicate data from the Ephnogram, FPCGDB, Pascal, PhysioNet16 and
		PhysioNet22 datasets respectively. The SSL and fully-supervised baseline
		models are trained and evaluated on \textit{In-Distribution} data from the
		respective dataset. The presented models are the best-performing models, based on results from the \textit{In-Distribution} test data split. Once trained, they are then evaluated on
		out-of-distribution data originating from unseen datasets. The `All' labels
		indicate that the PCG signals are classified based on the dataset's original
		labels (P16 has only binary labels), whereas `Binary' indicates that the
		signals are classified between normal and abnormal.  The SSL-NoDs model
		refers to the backbone network which has not `seen' data from the downstream
		task dataset during pretraining. 
	}
	\label{tab:experimental_results_ood}
	{
		\begin{tabular}{cc|cccc|cc|cc}
			\toprule
			~ & ~ & \multicolumn{4}{c|}{In-Distribution} & \multicolumn{2}{c|}{Out-of-Distribution} & \multicolumn{2}{c}{Out-of-Distribution} \\			
			~ & ~ & \multicolumn{2}{c}{All} & \multicolumn{2}{c|}{Binary} & \multicolumn{2}{c|}{Binary} & \multicolumn{2}{c}{Binary}  \\
			\textbf{SSL Datasets} & \textbf{Model} & \textbf{Acc} & \textbf{F1} & \textbf{Acc} & \textbf{F1} & \textbf{Acc} & \textbf{F1} & \textbf{Acc} & \textbf{F1} \\
			\midrule
			~ & ~ & \multicolumn{4}{c|}{\textit{Pascal}}& \multicolumn{2}{c|}{\textit{PhysioNet2016}}  & \multicolumn{2}{c}{\textit{PhysioNet2022}} \\
			--- & Baseline & 79.37 & \underline{48.37} & 67.77 & 40.00 & 51.95 & 62.74 & 51.33 & 59.41 \\			
			E, F, Pa, P16 & SSL & 76.87 & 42.18 & 31.25 & \textbf{47.61} & 63.55 & 74.88 & 53.24 & 64.44 \\			
			E, F, Pa, P22 & SSL
			& 76.87   &  42.18  & 32.10  &  \underline{42.25}
			&  61.56 &  72.69  & 61.11  & \underline{73.64}     \\			
			E, F, P16, P22 & SSL-NoDs
			&  75.62  &   39.06  &  31.25   & 40.00  
			&  59.88   &  \underline{74.97}  &  58.33  &  72.06  \\		
			all & SSL 
			& 81.25  &  \textbf{53.12}  &  35.93  &  \textbf{47.61} 
			& 67.13  &  \textbf{79.09}  &  62.11  &  \textbf{74.72} \\			
			\midrule
			~ & ~ & \multicolumn{4}{c|}{\textit{PhysioNet2016}}& \multicolumn{2}{c|}{\textit{Pascal}}  & \multicolumn{2}{c}{\textit{PhysioNet2022}} \\
			--- & Baseline 
			&  ---   &   ---   &  88.64  &  \textbf{93.09}   
			& 55.00  &  40.00  &  49.85  &  60.59  \\   
			E, F, Pa, P16 & SSL
			& ---  &  ---  &  80.07  &  87.38
			& 48.58  &  42.90  & 57.74 &  70.97 \\    
			E, F, Pa, P22 & SSL-NoDs
			&  ---   &   ---  &  74.97    &  85.69	  
			& 52.55  &  42.61   & 58.99 & 71.59 \\    
			E, F, P16, P22 & SSL
			&  ---   &   ---  &    84.19  &  89.65
			&  31.25  &   \underline{47.62}  &  62.07 &  \underline{78.07} \\    
			all & SSL 
			&  ---   &   ---   &  85.02  &  \underline{90.31}  
			& 42.19  &  \textbf{50.66}  &  70.45  &  \textbf{81.63} \\
			\midrule
			~ & ~ & \multicolumn{4}{c|}{\textit{PhysioNet2022}}& \multicolumn{2}{c|}{\textit{Pascal}}  & \multicolumn{2}{c}{\textit{PhysioNet2016}} \\    
			--- & Baseline
			&  90.80  &  \textbf{86.21}  &  86.90   &  \textbf{91.90}   
			&  53.12  &  \underline{53.12}  &  67.91   &   80.64    \\   
			E, F, Pa, P16 & SSL-NoDs
			&  87.94  &  81.91  &  82.49  &  89.71
			& 48.29  &  47.09  & 70.45 & 81.63\\    
			E, F, Pa, P22 & SSL 
			&  88.55   &  82.83   &   82.83  &   89.96  
			&  51.42   &   48.64  &  71.07  &  \underline{82.94} \\   
			E, F, P16, P22 & SSL
			&  85.37  &  78.06  &  78.06   &  87.67
			&  31.25 &  47.61  &  74.97 &  \textbf{85.70} \\    
			all & SSL 
			& 89.31  &  \underline{83.96}  &  83.67  &  \underline{90.00}  
			& 48.44  &  \textbf{54.79}   &  74.97  &  \textbf{85.70} \\    
			\bottomrule
		\end{tabular}
	}
\end{table*}

Table \ref{tab:experimental_results_ood} presents the Accuracy and F1-Scores of
the best performing models trained via contrastive SSL for all downstream
tasks, including evaluation on OOD distribution data, as presented in Section
\ref{eval_framework}. We perform a total of 4 cycles of SSL experiments,
evaluating all augmentation combinations in each cycle. In each experiment
cycle, we either use all available datasets or leave one labeled dataset out at
a time. The first column of Table \ref{tab:experimental_results_ood} denotes
the datasets used during SSL training. In the same table, the
`\textit{Baseline}' model refers to the fully-supervised model trained solely
on data from the downstream `In-Distribution' dataset. Due to the variance in the OOD results for the baseline model, we present the average of 5 total runs. Finally, the `SSL-NoDs' model refers to the model which has not 
been pretrained with data originating from the dataset of the downstream task. All results are rated based on the
F1-Scores on the `In-Distribution' test splits and the adopted augmentations for each case are the ones mentioned in Section \ref*{sec:aug_eval}. The same 
models are then evaluated on the test splits of the OOD datasets.

\subsection{Discussion}\label{sec:discussion}
At a first glance, the benefits of SSL pretraining admittedly do not seem very 
compelling. When evaluating the model on the test data of the In-Distribution 
data, the baseline fully-supervised model outperforms the SSL models in 2 out of 3 cases. When trained on the PhysioNet2016 and PhysioNet2022 datasets the 
baseline surpasses the best SSL model by 2.78\% and 1.9\% respectively. However,
the same does not hold in the case of the Pascal dataset, where the SSL model is
able to outperform the baseline by 4.75\% when evaluated on `All' labels and by 
7.61\% in the `Binary' task. Since the Pascal dataset is 
considerably smaller than the other 2, and arguably the hardest among the datasets, the above results validate the 
effectiveness of pretraining a model on a large corpus of data, when available 
labeled data is not sufficient.  

By diving further into the results, the true benefits of the proposed method
begin to show. First of all, the drop in effectiveness between in-distribution
and OOD data is apparent. Additionally, in all cases the baseline models appear to have overfit their training data distribution and fail to maintain their classification ability across distinct datasets. On the other hand, the models which have been pretrained via SSL yield far superior results. In all cases, the models trained with the proposed framework surpass the baseline by 11.68\% on average and even outperform it by 21.04\% in the case of the PhysioNet22 OOD task. Even more so, the SSL models were surprisingly able to surpass both the effectiveness of the baseline and their `In-Distribution' counterpart, in the case of the Pascal OOD evaluation. Specifically, the downstream models trained on data from PhysioNet2016 and PhysioNet2022 were able to yield better results than the fully-supervised model trained solely on Pascal data. Furthermore, even though there is a decrease in effectiveness when the OOD datasets are left out during model pretraining, the SSL models continue to outshine the baseline across the board, with the exception of the case where the downstream task is the PhysioNet2022 data and models are evaluated on OOD data from Pascal. Finally, we would like to note that the above findings are aligned with
the results of our preliminary SSL method \cite{ballas_listen2yourheart_2022}
submitted in the PhysioNet2022 challenge, where the proposed model was able to
avoid overfitting on the available training distribution and demonstrated
adequate generalization effectiveness on completely hidden data.

To further demonstrate the generalization capabilities of the SSL models, we 
provide t-SNE visualizations and illustrate the feature distribution of 
representations extracted from the penultimate layer of the best-performing 
model, in addition to representations extracted from a fully-supervised 
baseline model. Specifically, we visualize the features from signals drawn 
from distributions the models have been trained on (In-Distribution) and 
unseen distributions (OOD) in Fig. \ref{fig:tsne}. As depicted, the feature 
distribution of the baseline models are disordered, without a clear 
distinction between the abnormal (blue) and normal (signals). On the other 
hand, the features of the SSL model portray a somewhat clearer structure with
the features of each signal type grouping into clusters towards the upper-right and bottom-left corners. The compelling aspect of the visualization results, lies in the fact that the general feature distributions of the SSL models align on both in-distribution and completely unseen during training OOD samples.

\section{Conclusions \& Future Work}
\label{sec:conclusions}

This work provides a thorough evaluation of augmentation techniques and transformations for pretraining neural networks to achieve robust phonocardiogram (PCG) classification. Our method focuses on identifying the most effective augmentations for improving the performance of self-supervised learning (SSL) models. We present both the top-performing combinations of augmentations and a summary of the most frequently used transformation functions in the top 75 experimental results. Our findings highlight that the use of low-pass cutoff filters, consistently leads to the best-performing models across various datasets, making them essential for this task. Additionally, adding small random noise and applying signal inversion or reversal further enhance model performance.

Our results validate that contrastive SSL significantly improves the robustness and generalizability of PCG classifiers. Compared to fully-supervised models, SSL models experience a much smaller drop in effectiveness when evaluated on unseen data. However, the proposed method has limitations: SSL pretraining requires access to a large data corpus to maximize the model’s representation learning capabilities, and the contrastive loss function relies on large batch sizes, imposing a computational constraint.

In future work, we plan to refine the augmentation techniques by introducing new functions and exploring additional parameters, extend our approach to transfer learning tasks, and investigate the most effective augmentation policies for other 1D biosignals such as ECG, EMG, and EEG classification.

\bibliographystyle{IEEEtran}
\bibliography{bibliography}

\begin{IEEEbiography}[{
		\includegraphics[width=1in,height=1.25in,clip,keepaspectratio]{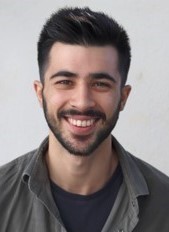}
      }]{Aristotelis Ballas}{\space} is currently working toward the
      Ph.D. degree in computer science with the Department of Informatics and
      Telematics, Harokopio University of Athens, Greece. His research interests
      include machine learning and algorithms for robust representation
      learning, with an emphasis on domain generalization and AI in healthcare.
\end{IEEEbiography}

\begin{IEEEbiography}[
  {\includegraphics[width=1in,height=1.25in,clip,keepaspectratio]{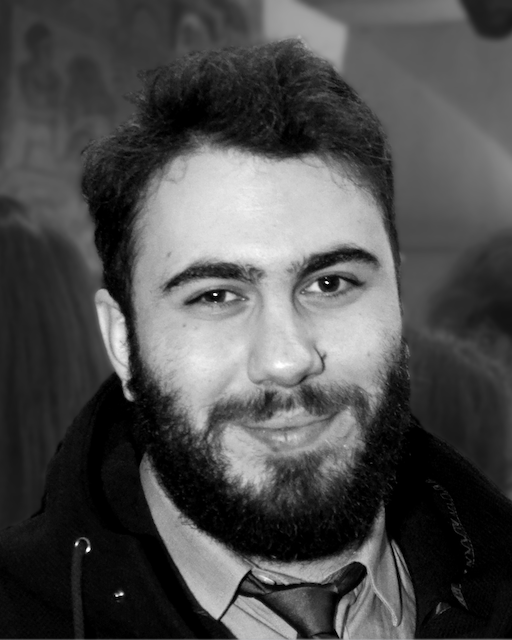}
  }]{Vasileios Papapanagiotou}{\space}
  \textit{Member}, IEEE, is an Assistant Professor at the Department of
  Medicine, Huddinge of Karolinska Institutet in Stockholm, Sweden. He
  received his Diploma and Ph.D. of Electrical and Computer Engineering in 2013
  and 2019 respectively, both from the Department of Electrical and Computer
  Engineering of Aristotle University of Thessaloniki in Greece. Since 2013 he
  has been mainly working as a researcher and on EU-funded research projects
  (FP7, H2020). His research interests include signal processing and machine
  learning with applications to wearable devices and sensors in the context of
  behavioral monitoring and quantification. He is a member of IEEE, IEEE EMBS,
  and the Technical Chamber of Greece.
\end{IEEEbiography}

\begin{IEEEbiography}[
	{\includegraphics[width=1in,height=1.25in,clip,keepaspectratio]{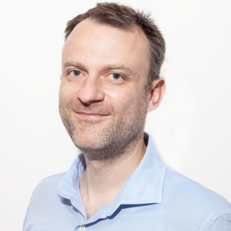}
	}]{Christos Diou}{\space}
	\textit{Member}, IEEE, is an Assistant Professor of Artificial Intelligence
	and Machine Learning at the Department of Informatics and Telematics,
	Harokopio University of Athens. He received his Diploma in Electrical and
	Computer Engineering and his PhD in Analysis of Multimedia with Machine
	Learning from the Aristotle University of Thessaloniki. He has co-authored
	over 80 publications in international scientific journals and conferences and
	is the co-inventor in 1 patent. His recent research interests include robust
	machine learning algorithms that generalize well, the interpretability of
	machine learning models, as well as the development of machine learning models
	for the estimation of causal effects from observational data. He has over 15
	years of experience participating and leading European and national research
	projects, focusing on applications of artificial intelligence in healthcare.
\end{IEEEbiography}


\EOD

\end{document}